\documentclass[letterpaper, 10 pt, conference]{ieeeconf}  % Comment this line out if you need a4paper
\usepackage{graphicx}

\IEEEoverridecommandlockouts                              % This command is only needed if 
\usepackage{url}
\usepackage{graphics} % for pdf, bitmapped graphics files

\usepackage{xcolor}
\usepackage{CJKutf8}
\definecolor{awesome}{rgb}{1.0, 0.13, 0.32}

\title{\LARGE \bf
Cooking Task Planning using LLM and Verified by Graph Network
}

\author{Ryunosuke Takebayashi$^{1}$, Vitor Hideyo Isume$^{1}$, Takuya Kiyokawa$^{1}$, Weiwei Wan$^{1}$ and Kensuke Harada$^{1,2}$% <-this % stops a space
\thanks{This work was supported by the New Energy and Industrial Technology Development Organization (NEDO) under Project JPNP20006}
\thanks{$^{1}$All the authors are with Graduate School of Engineering Science, The University of Osaka, 1-3 Machikaneyama, Toyonaka 560-8531, Japan
        {\tt\small vitor@hlab.sys.es.osaka-u.ac.jp}}%
\thanks{$^{2}$Kensuke Harada is with National Inst. of Advanced Industrial Science and Technology,
        Tokyo, Japan}%
}

\begin{document}

\maketitle
\thispagestyle{empty}
\pagestyle{empty}

%%%%%%%%%%%%%%%%%%%%%%%%%%%%%%%%%%%%%%%%%%%%%%%%%%%%%%%%%%%%%%%%%%%%%%%%%%%%%%%%
\begin{abstract}

%Newest
Cooking tasks remain a challenging problem for robotics due to their complexity. Videos of people cooking are a valuable source of information for such task, but introduces a lot of variability in terms of how to translate this data to a robotic environment. This research aims to streamline this process, focusing on the task plan generation step, by using a Large Language Model (LLM)-based Task and Motion Planning (TAMP) framework to autonomously generate cooking task plans from videos with subtitles, and execute them. Conventional LLM-based task planning methods are not well-suited for interpreting the cooking video data due to uncertainty in the videos, and the risk of hallucination in its output. To address both of these problems, we explore using LLMs in combination with Functional Object-Oriented Networks (FOON), to validate the plan and provide feedback in case of failure. This combination can generate task sequences with manipulation motions that are logically correct and executable by a robot. We compare the execution of the generated plans for 5 cooking recipes from our approach against the plans generated by a few-shot LLM-only approach for a dual-arm robot setup. It could successfully execute 4 of the plans generated by our approach, whereas only 1 of the plans generated by solely using the LLM could be executed.

% Old English Version
%In this study, we propose a Large Language Model (LLM)-based Task and Motion Planning framework that enables a robot to autonomously generate and execute a cooking task plan from cooking videos with subtitles. Conventional LLM-based task planning methods are not generally reliable for this task for two reasons: the uncertainties in the video, and the hallucinations from the LLM. To address these problems, this study explores the use of LLMs in combination with a Functional Object-Oriented Network (FOON). Using this combination, it is possible to generate task sequences with manipulation motions that are both logically correct and executable by robots. The experiments show that our method is capable of generating logically correct motions, whereas using a few-shot LLM-only approach would typically fail.

% Original text
%本研究では, 字幕付き料理動画を入力とし, 自律的にロボットが調理作業計画を作成・実行できる，LLMベースのTask and Motion Planningのフレームワークを提案する. 
%Conventional LLM-based task planning is not always reliable for two reasons: one is the uncertainties existing in the video image, and the other is the hallucination of LLM. 
%To address this problem, this study explores the use of large language models (LLMs) in combination with a Functional Object-Oriented Network (FOON). With this combination, it is possible to generate task sequences with manipulation motions that are both logically correct and executable by robots. The results of the experiments confirm the logical correctness of the generated motion, whereas the method without the proposed approach could not.

\end{abstract}

%%%%%%%%%%%%%%%%%%%%%%%%%%%%%%%%%%%%%%%%%%%%%%%%%%%%%%%%%%%%%%%%%%%%%%%%%%%%%%%%
\section{Introduction}

% Newest
With the widespread popularity of video streaming platforms, cooking recipes are increasingly being shared worldwide through detailed online video demonstrations (hereby referred to as "cooking videos"). Generating cooking task plans suited for a robot environment from such videos could greatly improve their versatility and possible practical applications. A key issue of using these videos, however, is their non-standardized structure. Specifically, they may vary in camera angles and tools used, which can hinder accurate recognition of the state of the scene. In addition, there may be differences in the space arrangement shown in the video and the space arrangement for the robot environment. Further, many videos edit out certain actions, such as preparing or moving kitchen utensils, so the robot system must also infer and compensate for these omitted steps.

To overcome these challenges, we propose a novel method that combines a multimodal Large Language Model (LLM) with Functional Object-Oriented Network (FOON)\cite{FOON1}\cite{FOON2}\cite{FOON3}\cite{FOON4}, a graph-based representation of task structures. LLMs possess general knowledge and reasoning capabilities learned from diverse datasets, enabling them to interpret and understand images from cooking videos combined with language instructions to generate corresponding robot actions. However, LLMs may produce factually incorrect outputs, due to hallucination or not following the prompts, which can be more prominent in complex, long-horizon tasks. Therefore, to ensure planning accuracy, we integrate the task graph structure from FOON. In our proposal, FOON represent cooking tasks as graph structures, allowing for environment-independent task planning. By comparing the inputs and outputs of FOON's functional units, we can detect logical inconsistencies in the plan at each step, and generate feedback prompts to correct them accordingly.

\begin{figure}[tbp]
  \begin{center}
    \includegraphics[width=\linewidth]{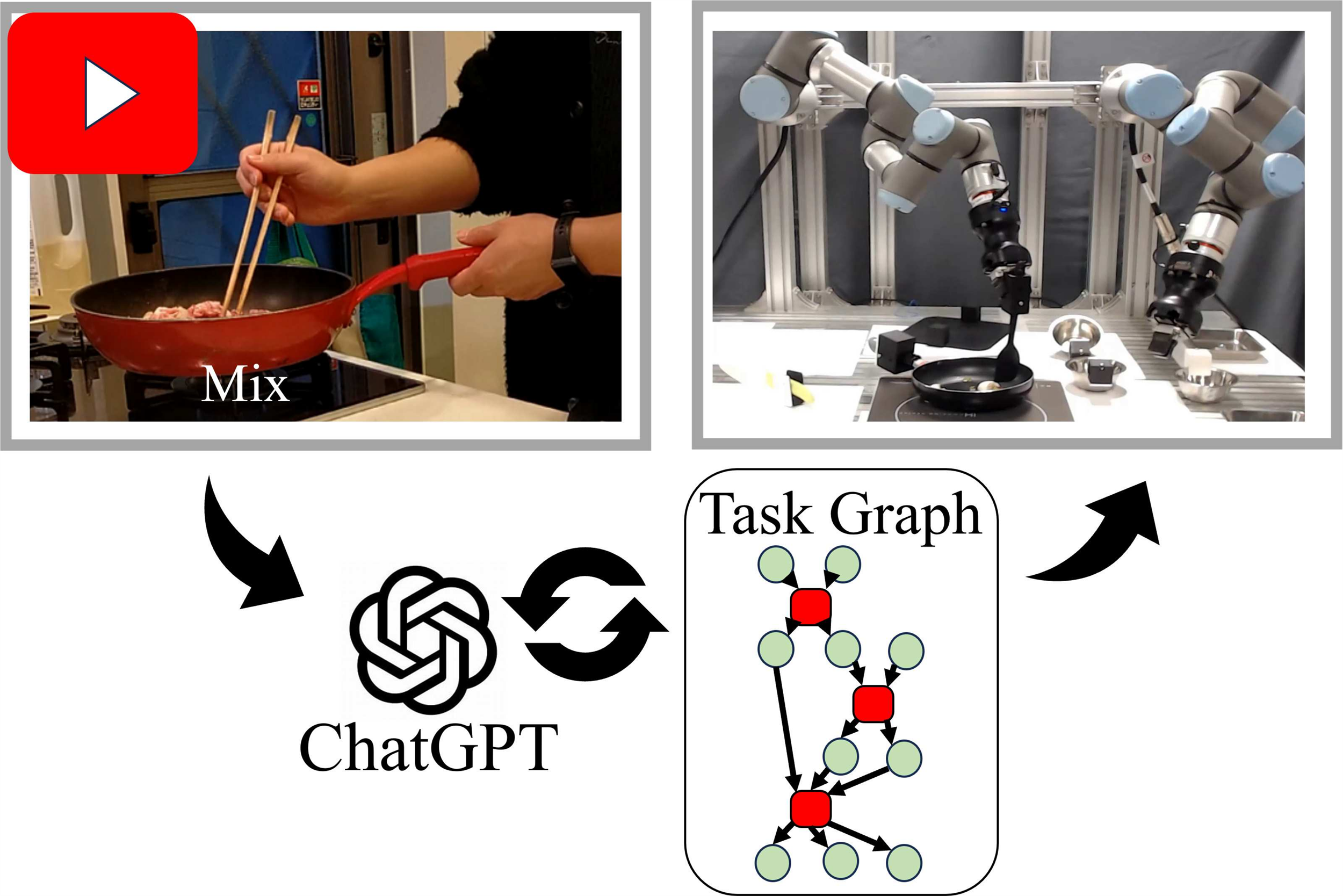}
    \caption{Our proposed framework to generate a cooking task plan for a robot from a video, using LLMs for reasoning and a task graph structure for logical validation.}
    \label{fig:intro}
  \end{center}
\end{figure}

In this work, we first use an LLM to infer the preparation states of the target food throughout the scenes of a cooking video and to generate a set of cooking tasks plans. Each plan is, then, converted into a functional unit for FOON, where its feasibility in the local environment can be verified. If the plan is deemed infeasible, FOON identifies the likely cause, generating appropriate feedback for the LLM to revise this plan. Using this iterative process, FOON compensates for possible planning errors from the LLM, until all tasks plans for a recipe are successfully generated, producing an executable cooking task plan.

The main contribution of this work is the integration of an LLM and FOON in order to generate cooking task plans from a subtitled cooking video, with error correction and environment-independent planning. We evaluate the proposed framework in terms of information extracted from the video, such as the subtitles, object and ingredient states estimation, and the execution of five cooking recipes, where each recipe is a set of cooking tasks from a single video.

\section{RELATED WORKS}

%Newest
Traditionally, task planning for robotics has relied on methods based on PDDL \cite{PDDL1}\cite{PDDL2} or task graphs like FOON \cite{FOON1}\cite{FOON4}. While they can ensure logical consistency, they lack scalability and flexibility. 

Alternatively, some methods have opted to use neural networks to obtain task instructions directly from the input, such as LLM-based approaches \cite{chatGPTforrobotics}\cite{Erra}\cite{Text2motion}\cite{Progprompt}, vision-language models (VLMs) and other multimodal language models \cite{Lm-nav}\cite{SocraticModels}, and even diffusion-based for videos \cite{CLAD}. But, the possibility of hallucination in LLM-generated outputs \cite{hallucination} still presents a risk in regards to the accuracy of a generated task plan. To address this, some studies have used prompt-based feedback mechanisms \cite{Erra}\cite{Text2motion}\cite{Progprompt}. These approaches, however, do not verify the logical correctness of the plan.

To overcome these limitations, hybrid approaches combining LLMs with classical planning methods have been proposed \cite{LLMP}\cite{LLMFOON1}\cite{LLMFOON2}. Still, they do not effectively address cases where the LLMs fails to design PDDL domains due to hallucinations or other errors derived from its output.

To handle these challenges, this work explores an approach to leverage LLMs for long-horizon task planning. By converting LLM-generated task plans into task graphs, our method allows for validation checks and re-planning, ensuring logical consistency, being well-suited for complex task planning such as a cooking task.

\begin{figure}[tbp]
  \begin{center}
    \includegraphics[width=\linewidth]{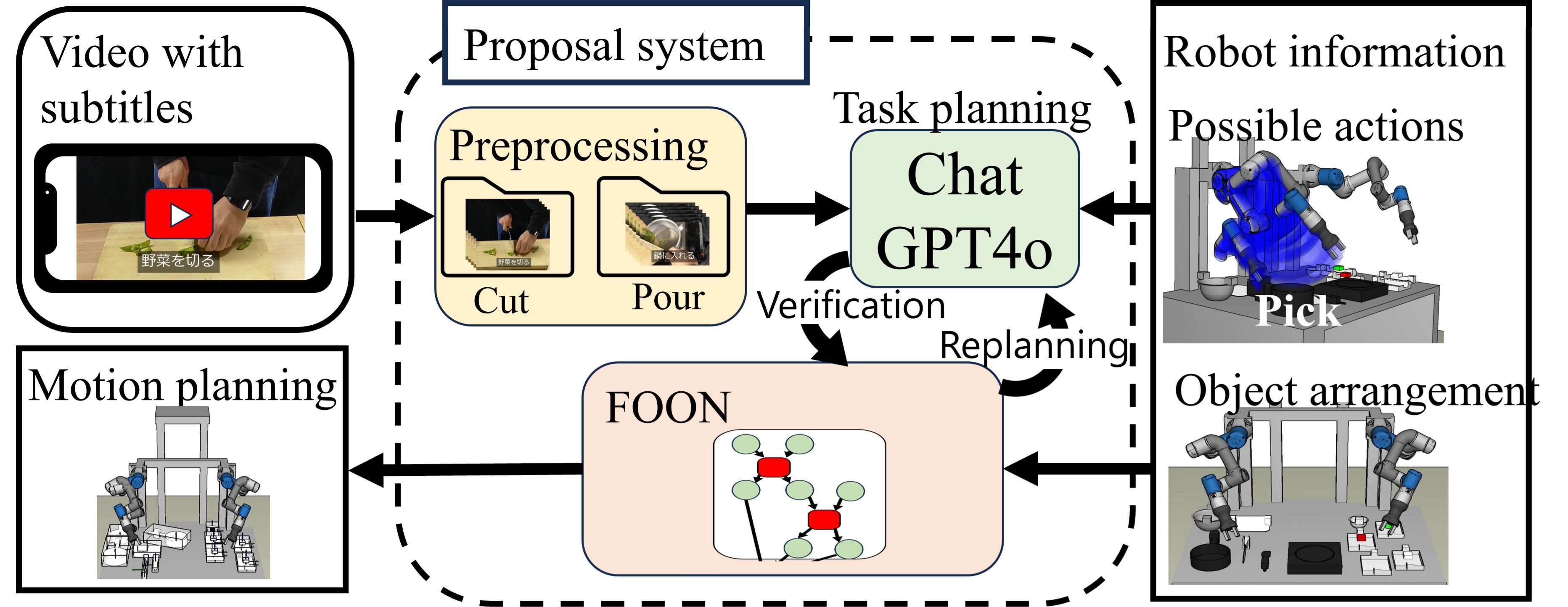}
    \caption{Overview of our proposed method.}
    %\caption{Overview of our proposed method \wan{図1と2のタイトルは同じ。両図とも第三章からしか引用していません。第1章のページに置くのは醜いです。}}
    \label{fig:overview}
  \end{center}
\end{figure}

\section{Proposed Method}

\subsection{Overview}

% Newest
Fig. \ref{fig:overview} shows the overview of the proposed method. From a Youtube video with subtitles of a human cooking, a series of images are extracted. These images, along with a set of possible robot actions and the current arrangement of the robot's workspace are provided as the input to the LLM, which generates the cooking task plans for the robots as a sequence of actions. Then, these plans are sequentially converted to a FOON's functional unit and added to the graph structure. If the conversion fails, the plan where the failure happened and the cause of failure are detected and provided to the LLM in an error prompt as feedback for re-planning. This process is iterated until the entire FOON is successfully generated, which is used for the robot's motion planning.

% Old English version
%Fig. \ref{fig:overview} shows the overview of the proposed method. We use a YouTube video of human cooking with subtitles. A series of images extracted from the video, the possible actions of the robot, and the current object arrangements of the robot's workspace are the inputs to the LLM, which generates the robot's action sequence. Then, we try to generate FOON from the action sequence.  Here, if the action sequence is not correctly generated, we detect the source of failure in the action sequence. Then LLM regenerates the action sequence by taking into account the error prompt. This process is iterated until FOON is successfully generated. Finally, the robot motions are planned from the generated FOON.  

\subsection{Functional Object-Oriented Network (FOON)}

%Newest
In our work, we use FOON, a representational form of task-sequence graph, to validate the generated cooking task plans from the LLM, and for object motion planning. In this section, we briefly explain its basic structure and our extension to the original object nodes from FOON. 

% Old English text
%In this study, we generate robot actions from FOON, a representational form of a task-sequence graph.
%We briefly explain its basic structure and our extension to the original FOON. 

\subsection{Graph structure}

\begin{figure}[tbp]
  \begin{center}
    \includegraphics[width=0.7\linewidth]{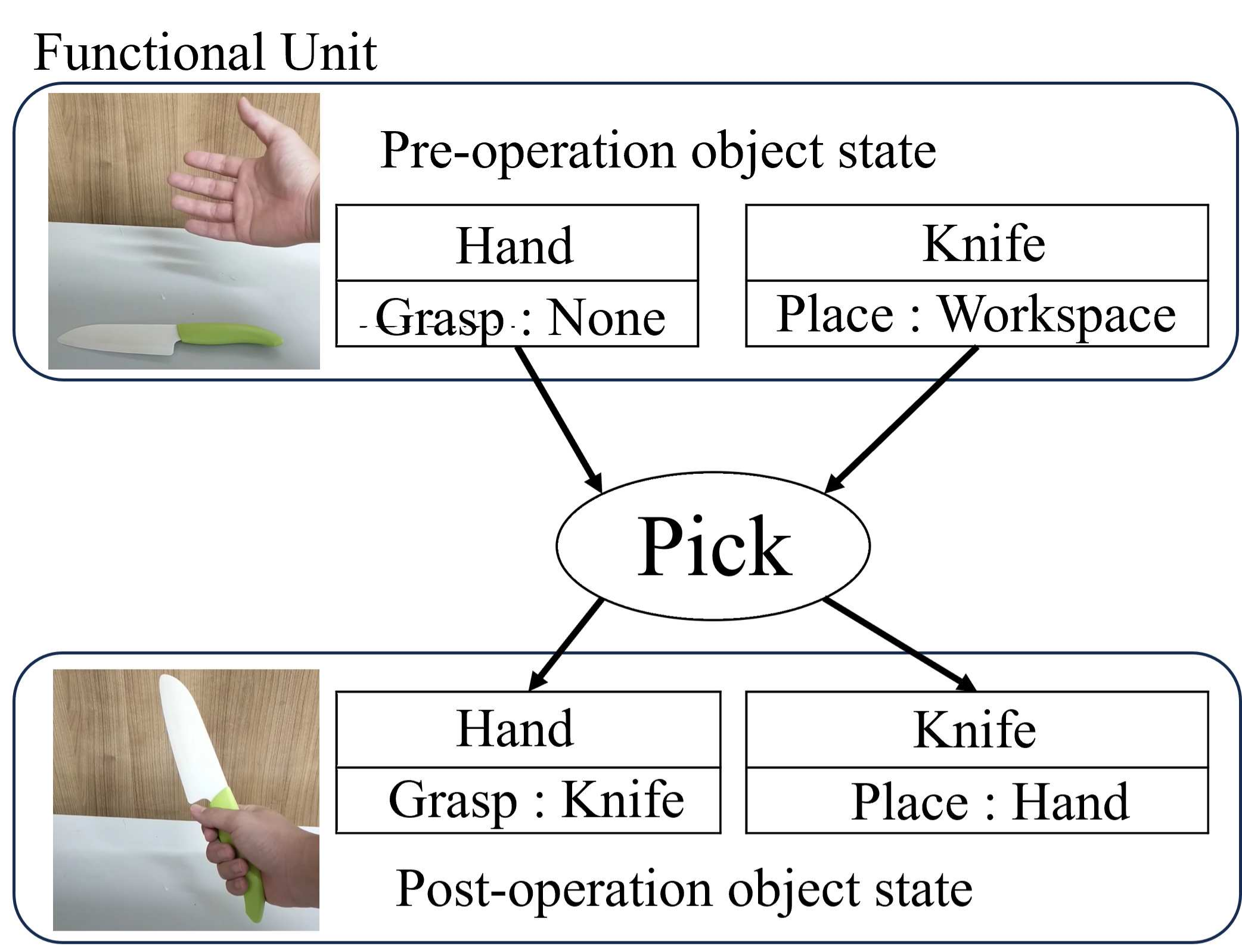}
    \caption{Example of a functional unit for a "Pick" motion.}
    \label{fig:Functional_Unit}
  \end{center}
\end{figure}

%Newest
An example of the FOON's smallest unit, called a \textit{functional unit}, is shown in Fig. \ref{fig:Functional_Unit}. In our work, we follow the functional unit structure from \cite{FOON4}:
\begin{itemize}
    \item \textbf{Object nodes}: representing ingredients, cooking utensils or containers;
    \item \textbf{Hand nodes}: representing the hand state;
    \item \textbf{Motion nodes}: description of the action;
\end{itemize}

Each functional unit represents a single action with input object and hand nodes, representing their state before the action, and output object and hand nodes, representing their state after the action described in the motion node. For example, in Fig. \ref{fig:Functional_Unit}, the input nodes for the "Pick" motion are the "hand" and the "knife". Initially, the hand is empty and the knife is placed on the workspace. Then, after the "Pick" motion is executed, the states change, with the hand now holding the knife, as shown in the output nodes. 

A task graph can be constructed by combining multiple functional units, and since each object node represents both the required object state before an action and after, it is possible to verify if the object states conditions are satisfied for an action in a step-by-step manner. This property will be leveraged to perform feasibility checks of the cooking task plans generated by the LMM, ensuring logical soundness in the overall cooking task.

\subsubsection{Environment object nodes and Target object nodes}
\begin{figure}[tbp]
  \begin{center}
    \includegraphics[width=\linewidth]{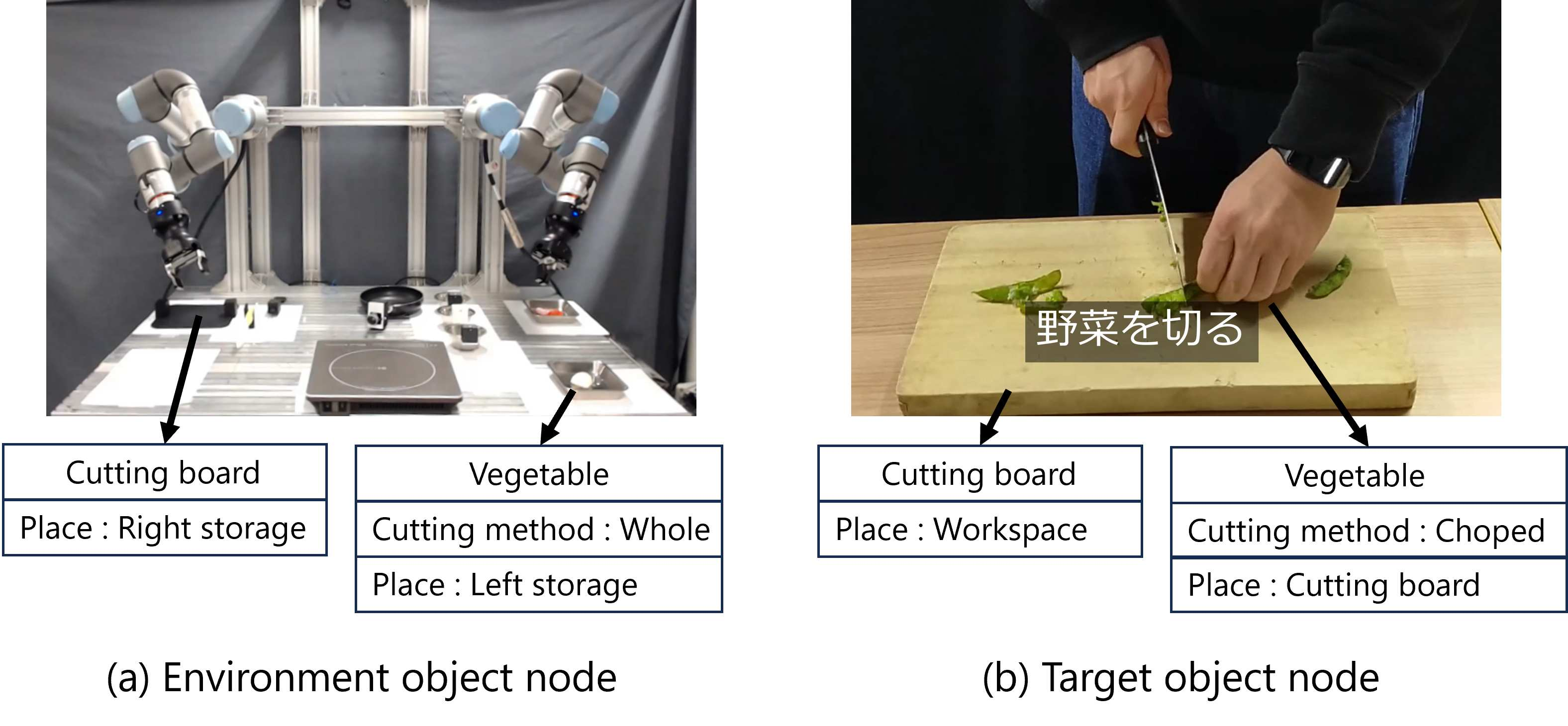}
    \caption{Additional types of Object nodes proposed for this task.}
    \label{fig:object_node}
  \end{center}
\end{figure}

%Newest
In this work, we define two additional types of object nodes, besides the original object node: \textit{environment object node} and \textit{target object node}, as illustrated in Fig. \ref{fig:object_node}. \textit{Environment object node} represent the state of all objects in the robot's local execution environment, with extra attributes such as the object placement relative to the environment (e.g. "Cutting board" is in the "Right storage", which is a predefined region of the local environment), and cooking status of ingredients. These nodes are used to keep track of changes in the state of objects in the robot's local environment. \textit{Target object node}, on the other hand, represent the estimated state of all objects observed in the reference video frames. It also has extra attributes, such as the cooking status attribute, and are used to define the desired object states in the output of a functional unit. To complete a cooking task successfully, the system needs to update the environment object node through multiple actions in order to achieve all states of the target object node.

%Old english text
%In this study, we define two additional types of object nodes, aside from the usual Functional Unit's object node: "Environment object nodes" and "Target object nodes", as illustrated in Fig. \ref{fig:object_node}. "Environment object nodes" represent the state of objects in the local execution environment, adding attributes such as the object placement and cooking status. Since they are updated through actions (Functional Units), they are used to manage the state of objects in the local environment. "Target object nodes", on the other hand, represent the state changes of objects observed in a reference video. By processing the video frames, the state of the objects are estimated, and the target food state that the robot needs to achieve is stored. Each processed frame has a corresponding Target object node. To complete the cooking task successfully, the system aims to update Environment object nodes through multiple actions in order to achieve the states of all Target object nodes.

%original text
%\subsubsection*{2)環境オブジェクトノードと目標オブジェクトノード}
%本研究では，機能ユニットのオブジェクトノードとは別に環境オブジェクトノードと目標オブジェクトノードを図\ref{fig:object_node}のように定義する．環境オブジェクトノードは想定環境にあるオブジェクトの状態を表し，オブジェクトの配置位置や調理状態を属性として持つ．そのため環境オブジェクトノードは操作(機能ユニット)により状態が更新され，環境のオブジェクト状態の管理に用いる．それに対して, 目標オブジェクトノードは，動画内のオブジェクトの状態変化を表現する．動画の前処理により作成した動画の画像から，オブジェクトがどのような状態にあるか推定し，ロボットにより達成すべき食材の状態を保存する．目標オブジェクトノードは前処理により作成した画像ごとに用意する．環境オブジェクノードを複数の操作により更新し，すべての目標オブジェクトノードの状態を達成していくことで料理の完成を目指す．

\subsubsection{Functional unit with variables}
\begin{figure}[tbp]
  \begin{center}
    \includegraphics[width=\linewidth]{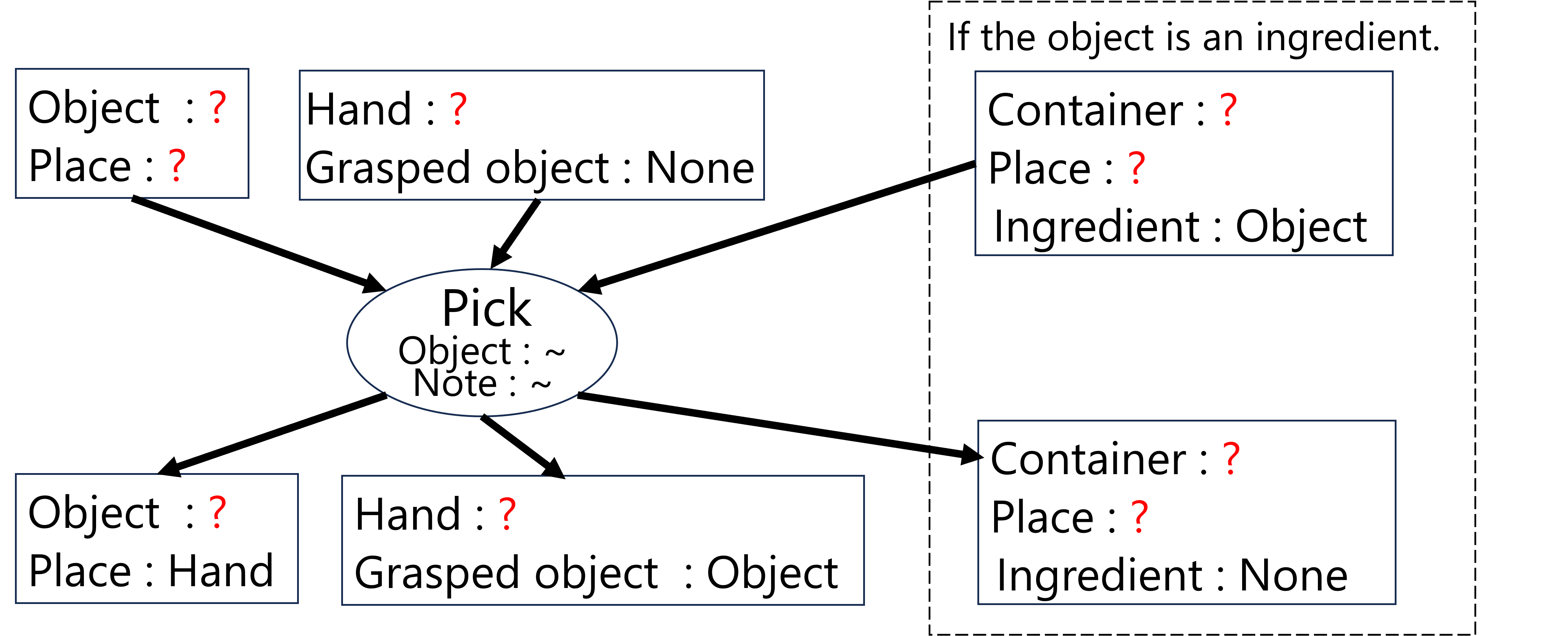}
    \caption{Functional unit with variables(Pick). The fields with the interrogation mark are filled with the values specified for the variables when instantiated.}
    \label{fig:Pick}
  \end{center}
\end{figure}

%New
In this work, we also use the \textit{Functional Unit with variables}, as proposed by Takada et al. \cite{FOON4}, to convert text-based task plans into graphs. This type of functional unit has a motion node, but object-dependent elements are treated as variables, acting akin to a template for possible variations of the same action. For example, in the case of a "Pick" motion, as shown in Fig. \ref{fig:Pick}, which hand is used and the object's "place" attribute may change depending on the object's location. However, a "Pick" motion will always result in the target object being held by the hand, allowing us to re-use this functional unit for the task graph with different objects, or objects in different locations.

%Old english version
%In this study, we use Functional Unit with variables, as proposed by Takada et al. \cite{FOON4}, to convert text-based task plans into graphs. Even for the same action, the structure of a Functional Unit can change depending on the target object and its location. For example, in the case of a "Pick" motion, as shown in Fig. \ref{fig:Pick}, which hand is used and the object's state may change depending on the object's location. However, a "Pick" motion will always result in the target object being held by the hand. To handle these variations, we use Functional Unit with variables, where object-dependent elements are treated as variables. Task graphs are systematically generated by assigning specific values to the variables for each task.
%\subsubsection*{3) 変数付き機能ユニット}
%本研究では高田ら\cite{FOON4}の変数付き機能ユニットを用いて，テキスト形式の作業計画をグラフに変換する．同じ動作であっても，対象物やその配置場所によって，機能ユニットの構造が変化する．例えばPick動作であれば，図\ref{fig:Pick}のように把持するオブジェクトの配置場所により，使用するハンドや対象物の状態が変化する．しかしながら，Pickモーションにより，把持するオブジェクトノードの状態がハンドに把持されている状態に変化する点は共通している．このような対象物によって変化する部分を変数として持つ変数付き機能ユニットを用意し，タスクごとに変数を代入することでグラフに変換する．

%%%%%%%%%%%%%%%%%%%
%竹林君確認
%%%%%%%%%%%%%%%%%%%%

%\section{Proposed Method}
%\wan{第2章：Method Overview -> この章: Proposed Method、なかなか変な名付けでした。}
\subsection{Video Pre-processing}

\begin{figure}[tbp]
  \begin{center}
    \includegraphics[width=\linewidth]{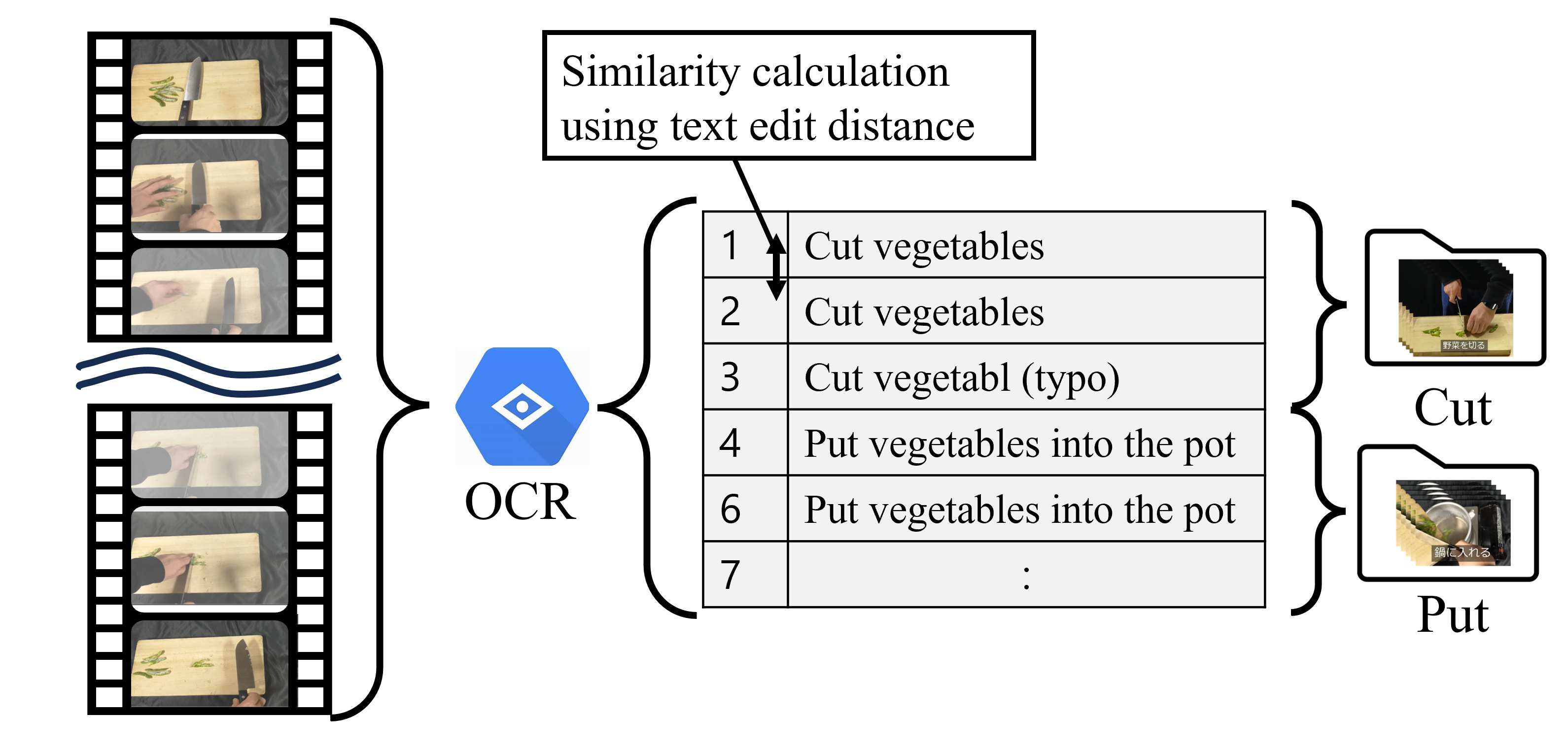}
    \caption{Video Pre-processing step. The subtitles are extracted using OCR, and the video is segmented according to text similarity with the possible actions.}
    \label{fig:preprocess}
  \end{center}
\end{figure}

% New
To generate the cooking task plans for the robot, we segment the video into distinct scenes. In addition, due to input restrictions in the LLM model used in this work, the video must first be converted into key frames, representative images from the video that capture key features of the cooking scenes. Therefore, we pre-process the video by extracting frames and segmenting scenes based on subtitle information obtained via optical character recognition (OCR), as illustrated in Fig. \ref{fig:preprocess}. To create a single image representing each segmented scene, the corresponding key frames are arranged in a 3×3 grid, serving as the input image for the LLM.

%Old
%To generate food cooking motions for a robot, we segment the video into distinct scenes. Additionally, since the ChatGPT model used in this study doesn't support video input, it must first be converted into representative images that capture key features of the cooking scenes. Therefore, in this study, we preprocess the video by extracting frames and segmenting scenes based on subtitle information obtained via optical character recognition (OCR), as illustrated in Fig. \ref{fig:preprocess}. Furthermore, to create a single image representing the entire video, we arrange the segmented images in a 3×3 grid, which serves as the input image for the large language model (LLM).
%ロボットが調理作業を行う場合，動画をシーンごとに分割する必要がある．また，本研究で使用するChatGPTのモデルは画像入力のみ可能なので，動画を調理シーンの特徴を表す画像に変換する必要がある．そこで本研究では，図\ref{fig:動画の前処理}のように動画をフレームごとに分割し，画像内の字幕情報を光学的文字認識により読み取ることで字幕ごとにシーン分割する．また，分割した画像を3×3で並べることで，動画を表す1枚の画像を作成し，LLMへの入力画像とする．

\subsection{Graph Generation}

In this step, we generate the graph of robot tasks from the preprocessed video images.

\subsubsection{Estimation of Target Object Node}
%New
For a cooking task plan, it is crucial to accurately determine the state of objects, such as ingredients, that are affected by the cooking actions. Here, we use the LLM to infer the target object node by prompting it with the background information, task description, thought process, constraints, allowed actions, attributes of the object node and subtitles alongside the images extracted from the video. This prompt is illustrated in  Fig. \ref{fig:Prompt}(a). 
To make the structure of the information clearer for the LLM, the type of information is categorized into "System" and "User" roles. In the system role, the overall task information, desired format and reasoning process is laid out. We provide background knowledge of FOON and role-playing prompting to improve the response accuracy. The task description explicitly instructs the LLM to estimate the state of the target objects, while the thought process is outlined to improve inference by following a Chain of Thought (Cot) approach. Constraints are defined to ensure that the output is formatted as an executable graph in Python.

In the user role, information about the possible robot actions, the object node attributes, such as position and cooking state, and the subtitles and key frames of the video are given. Providing the LLM with these concrete details and conditions, we facilitate reasoning for the target object node estimation. Additionally, we use few-shot prompting to help guide the LLM to output the target object node and motion nodes. The LLM is instructed to output these results in JSON format, which are converted into nodes and edges according to our graph representation.

\subsubsection{Action Estimation}

\begin{figure*}[t] % * をつけると2段組みを超える
    \centering
    \includegraphics[width=\textwidth]{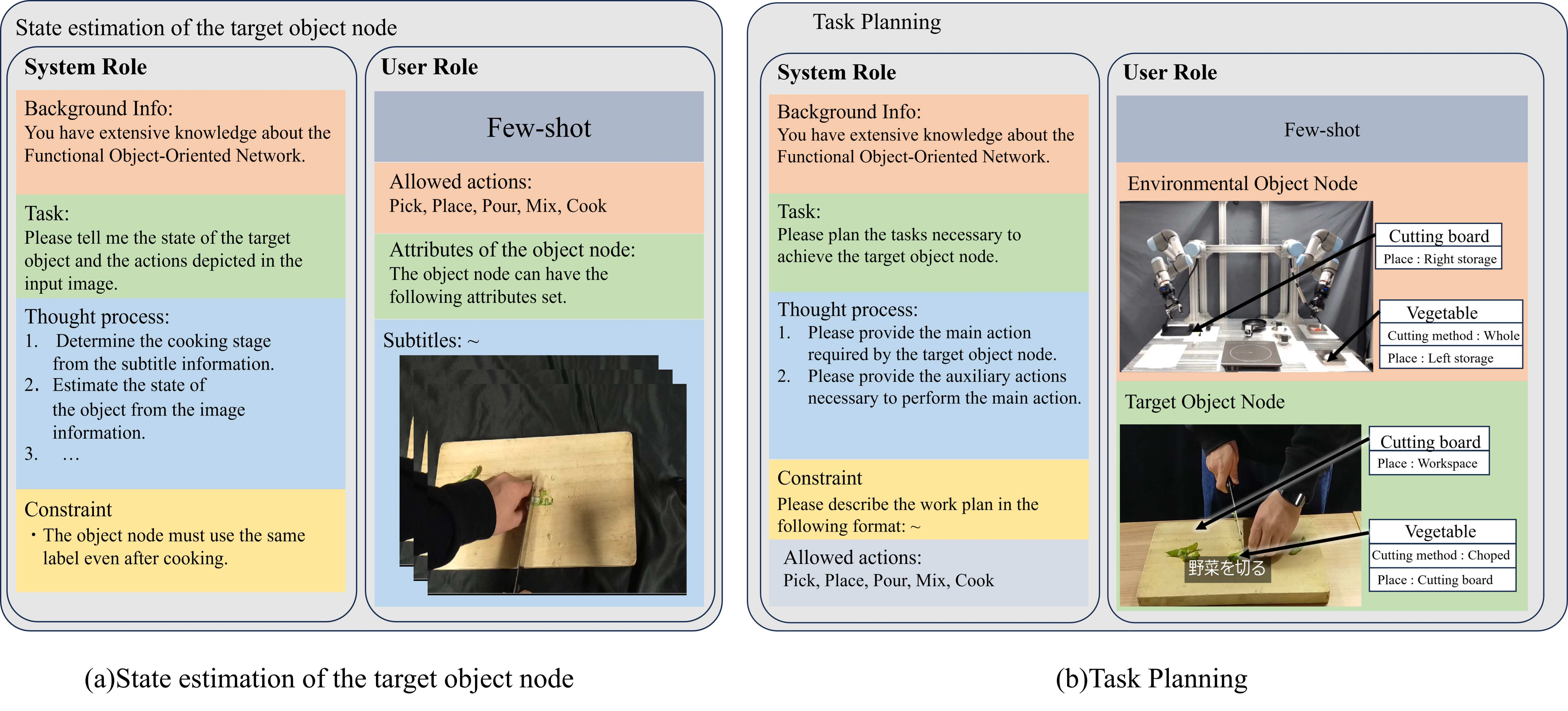} % 画像幅をページ全体に
    \caption{Overview of prompts used to generate task graph. The one in the left (a), is designed to estimate the required object states from the video, while the one in the right (b), is designed to plan how to achieve such states in the local environment.}
    \label{fig:Prompt}
\end{figure*}

%New
To complete a cooking task successfully, we need to determine the series of robot actions to achieve the estimated target object node state. Here, we use another LLM agent to compare the states of the objects observed in the video at the end of each task with the states of the objects in the local environment, and then, infer the necessary operations. The structure for the prompt for this step is illustrated in Fig. \ref{fig:Prompt}(b).

The system role follows a similar structure to the prompt used for target object node estimation, with modifications to the contents of task description, thought process and constraints. The user role provides the information of the environment object node and the target object node, both provided in JSON format. Using these prompts, the agent is responsible for inferring the sequence of allowed actions that transition the initial environment object node states to the target object node states. 

The output of the LLM consists of the name of the chosen action, and the corresponding object-dependent variables from the associated functional unit with variables, such as the object location and the hand used for the action.

% Old English text
%By sequentially executing robot operations to achieve the estimated target object states, the cooking process can be completed. To facilitate this, we compare the states of two types of object nodes: environment object nodes and target object nodes, and use an LLM to infer the necessary operations. The structure of this prompt is illustrated in Fig. \ref{fig:Prompt}(b).
%The system role follows a similar structure to the one used for target object node estimation, with modifications to the task content, reasoning method, and constraints. The input for the user role consists of the states of the environment object nodes and target object nodes, both provided in JSON format. Using this prompt, the LLM is tasked with inferring the necessary operations to transition the environment object node states into the target object node states.
%The output consists of textual descriptions of the required actions, specifying the object locations and the hand to be used for execution. By providing these inferred operations, action nodes are generated, ensuring a structured approach to achieve the desired cooking process.

%推定した目標オブジェクトノードの状態を順番にロボットによる操作で達成していくことで，最終的に料理を完成させることができる．そこで環境オブジェクトノードと目標オブジェクトノードの２種類のオブジェクトノードの状態を比較し，必要な操作をLLMにより推測する．プロンプトの構造を図\ref{fig:Prompt}(b)に示す． システムロールは, 目標オブジェクトノードの推定と同様な構造をしており，タスク内容や思考方法, 制約に変更を加えている．ユーザーロールへの入力内容は，環境オブジェクトノードの状態と,目標オブジェクトノードの状態をjson形式で与える．このプロンプトにより，環境オブジェクトノー ドの状態を目標オブジェクトノードの状態に変更するために必要なタスクをLLMに推定させる．出力では動作とその対象となるオブジェクトの配置場所や使用ハンドがテキスト形式で出力される．以上の方法により必要な操作を補完することで、アクションノードを生成する．

\subsubsection{Task validation and graph generation}
\begin{figure}[tbp]
  \begin{center}
    \includegraphics[width=\linewidth]{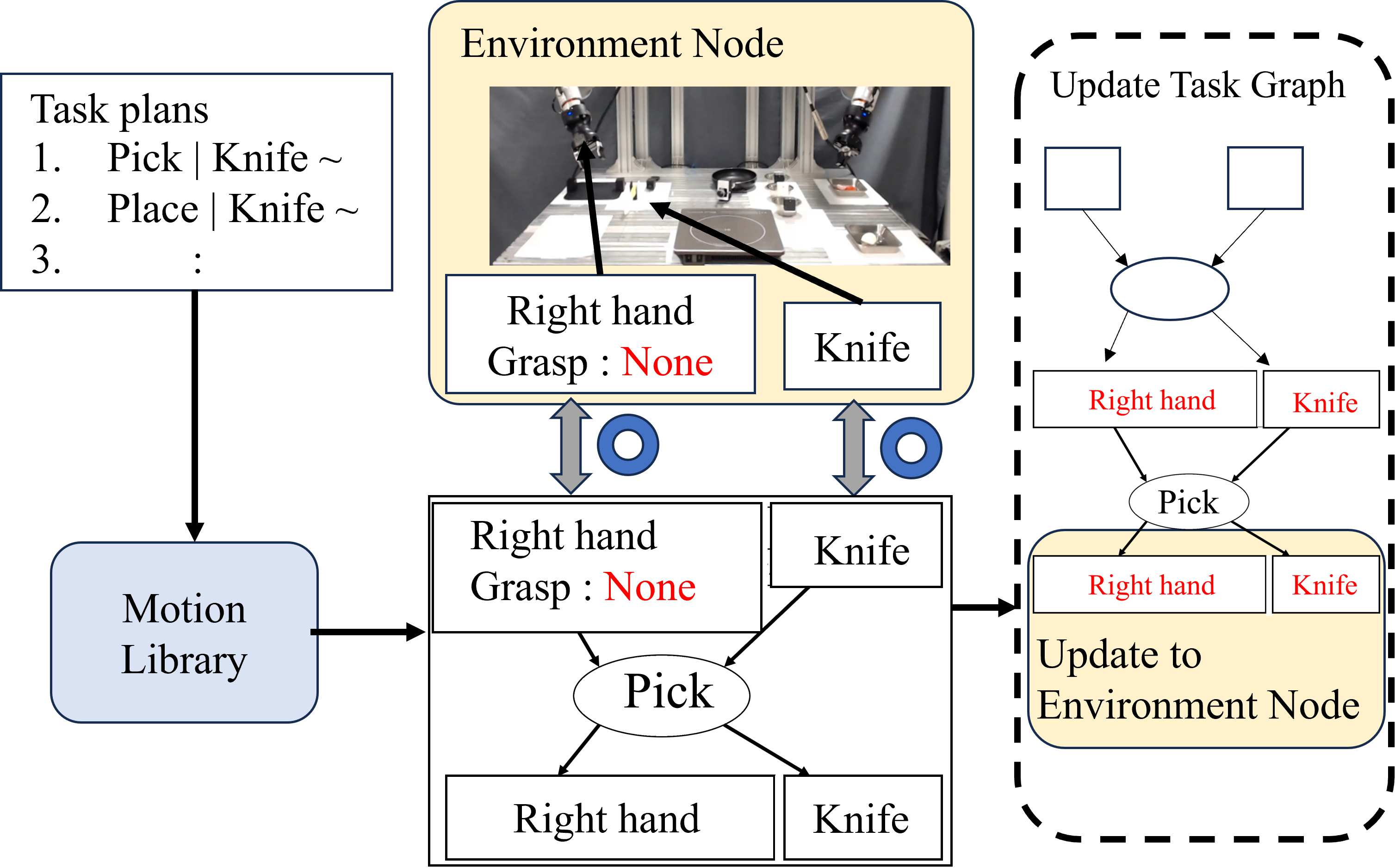}
    \caption{Task plan validation and graph generation}
    \label{fig:Task validation and graph generation:}
  \end{center}
\end{figure}

%New
Before adding the predicted motion node to to the task graph, the predicted variables need to be validated according to the state of the environment object node. The validation procedure is shown in Fig. \ref{fig:Task validation and graph generation:}, where a task plan predicted by the LLM, for example, the "Pick $|$ Knife $|$ Right hand $|$ Right storage", indicates that it should load, from a Motion Library, the functional unit with variables corresponding to the "Pick" motion and assign "Knife", "Right hand" and "Right storage" to the object name variable, the hand variable and the location variable, respectively.

Next, the inputs of the generated functional unit are compared with the states in the environment object node. As these inputs are the required conditions for the action, if their states match, the action is deemed feasible. Otherwise, if the attributes do not match, the task is considered infeasible, and re-planning is required. Once a task plan is validated, the environment object node is updated with the attributes of the outputs of the functional unit.

The text output generated by the LLM may contain errors caused by hallucinations. Therefore, when generating the functional unit to add to the task plan graph, it is necessary to validate the output. The process is repeated for all actions in the tasks plans, ensuring that a logically valid sequence of operations is generated.

\subsubsection{Replanning}
% new
In the case an error occurs during the task plan validation, the cause of the failure can be easily identified based on mismatched attributes between the environment node and the predicted functional unit variables. For example, in the "Pick $|$ Knife $|$ Right hand" task, the action requires the "Right hand" to be empty before grasping the object, if in the environment object node, the "Right hand" was already grasping another object, the task would be considered infeasible. This error is then provided to the LLM to generate a corrective action. This iterative process continues until all task plans are considered feasible. 

% old
%If errors occur in the task plan generated by the LLM, replanning is required. In this study, the LLM is utilized to infer the cause of task execution failure, and this inferred cause is incorporated into the replanning process. The feasibility of a task is determined by comparing the attributes of environment object nodes with those of the input nodes in the functional units.
%If a task is deemed infeasible, the cause of failure can be inferred based on the mismatched attributes. For example, in the task "Pick knife," the action requires that the hand is empty before grasping the object. If the attributes of the environment object node indicate otherwise, the task is considered infeasible. Specifically, if the hand is already holding another object, the action cannot be executed, and the object must first be placed down.
%By identifying the cause of task failure and prompting the LLM to generate a corrective action, the replanning process accounts for execution errors. This iterative process continues until all tasks are validated as executable.
%LLMによる作業計画に誤りがあった場合，再計画を行う必要がある．本研究では，作業計画が実行不可能な原因をLLMに推論させ，その原因を再計画に用いる．本研究では環境オブジェクトノードと機能ユニットの入力ノードの属性を比較することで実行可否判定を行う．もし，実行不可能と判定された場合は，その一致しなかった属性から実行不可能な原因を推定することが可能である．例えば，”Pick knife”というタスクがあった場合，ハンドが何も把持していない状態を必要とする．それに対して，環境オブジェクトノードの属性が異なっていた場合実行不可能だと判断される．
%ハンドが何かを持っている場合は実行不可能であり，一度手に持っているものを置く必要がある．このように実行不可能な原因を推定し，解決策をLLMに出力させることで失敗の原因を考慮した再計画を実現する．これをすべてのタスクが実行可能と判定できるまで繰り返す．

% 通常のsubsectionに戻す場合
%\renewcommand{\thesubsection}{\thesection.\arabic{subsection}}
\subsection{Motion Planning}

%New
Although the task graph generated by the LLM provides a logically correct sequence of tasks to complete the whole cooking process, it does not validate the actual motion execution. Information such as the required grasping pose, the pose and location of the objects is not considered. In a cooking task, which is a long-horizon task involving multiple objects, in a fixed robot limited workspace, the placement can significantly influence subsequent actions.

For example, consider the task of placing pork into a frying pan, executed by a dual-arm robot system. If the frying pan is handled by the right arm while the pork is handled by the left arm, in the execution sequence, the right arm places the frying pan in the workspace before the left arm places pork into it. However, even if the right arm can position the frying pan at the desired coordinates using inverse kinematics (IK), there is no guarantee that the left arm can achieve a feasible pose to place the pork into the pan.

Because this work mainly focus on the task plan generation, to avoid such issues, the robot's reachable workspace is discretized and we find initial object placement coordinates and poses that allow all tasks to be executed. Additionally, for each task, feasible motions are generated using custom functions that use methods such as RRT-Connect and linear trajectory planning. These functions take the object coordinates and grasping pose as the input.

\section{Evaluation}
%\wan{この章と次の章を一つの章の二つの小節にした方が良い気がします}
\subsection{Experiment on Graph Generation}
%New
To evaluate our proposed method, we use 10 cooking videos with subtitles from YouTube Shorts
\cite{YouTube1},\cite{YouTube2}, \cite{YouTube3},\cite{YouTube4}, \cite{YouTube5}. 
The selected videos exclusively feature cooking actions, such as ""cutting", "mixing", "heating", and "pouring", and each video has a duration of less than one minute. The dataset includes videos with different framing styles, such as close-up shots focusing only on the hands and videos featuring the cook. The video content also varies, with some focus solely on the cooking process, while other feature the cook tasting the dish and providing commentary. The success rate is evaluated for the three key steps in our planning generation procedure: video pre-processing, target object node estimation and task-sequence graph. 

%To evaluate the effectiveness of the proposed method, we conducted an evaluation experiment using 10 captioned cooking videos from YouTube Shorts. The selected videos exclusively feature cooking actions such as "cutting," "mixing," "heating," and "pouring," and each video has a duration of less than one minute. Additionally, the dataset includes videos with different framing styles, such as close-up shots focusing only on hands and videos featuring the cook.
%The video content varies, including both videos that focus solely on cooking and those where the cook tastes the dish and provides commentary. The success rate is evaluated for three key processes: video preprocessing, target object node estimation, and task plan estimation.
%本研究の手法の有用性を評価するために，Youtubeショートの10個の字幕付き料理動画を用いて，評価実験を行った．料理動画は「切る」，「混ぜる」，「加熱する」，「注ぐ」などの調理動作のみで完結する動画を使用しており，１分以内の動画である．また，料理動画には手元だけを映したものや調理者が映るものなど構図の違う動画を含んでいる. 動画内容に関しては，調理のみのものと，調理者が実食しコメントする動画も含んでいる．動画の前処理と，目標オブジェクトノードの推定，作業計画推定の３つの推定成功率について評価実験を行った．

\subsubsection{Video Pre-processing}
% new
In this evaluation, we measure the extracted subtitles information from the videos using OCR. The results are presented in Table \ref{tab:subtitle_error}. Across the 10 cooking videos, the amount of extracted subtitles exceeded the groundtruth amount by 17\%, due to OCR misrecognition and other factors. This primarily resulted in redundant instructions, which can be filtered by the LLM, ensuring that the task planning process remains unaffected.

% old
%The experimental results are presented in Table \ref{tab:subtitle_error}. Across the 10 cooking videos, the amount of extracted subtitles exceeds the actual amount by 17\%, due to OCR misrecognition and other factors. This primarily results in redundant instructions, which can be removed by the LLM, ensuring that the task planning process remains unaffected.

%実験結果を表\ref{tab:subtitle_error}に示す．実験結果から10個の料理動画に対して, OCRのご認識などにより平均17\%多くの字幕が抽出された．抽出した字幕数が実際の数よりも多いので，これは重複する指示が入るのみでLLMにより重複の除去が可能である．

\begin{table}[htbp]
\centering
\caption{Error analysis of number of subtitles extracted from the video.}
\begin{tabular}{|c||c|}
\hline
& Total Amount \\ \hline \hline
Groundtruth subtitles (words) & 230  \\ \hline
Extracted using OCR (words) & 270  \\ \hline
Error (words) & 40   \\ \hline
Error(\%)& 17   \\ \hline
Std. deviation (words) & 5.0 \\ \hline
\end{tabular}
\label{tab:subtitle_error}
\end{table}

\subsubsection{Target Object Node Estimation}

%New
Evaluation for this step is conducted on 276 images extracted from the 10 selected cooking videos. 
Among these images, 134 of them contain scenes unrelated to cooking, such as introduction sequences or redundant cooking operations. During this step, the LLM needs to correctly determine that such images do not require target object node estimation. It also needs to follow the provided instructions and output data in JSON format that is readable by Python to ensure they can be converted into a task graph after. Finally, it needs to accurately predict the target object node states. These three aspects are evaluated using ChatGPT-4o as the LLM and reported in Table \ref{tab:estimationsuccess}.

Our results indicate that the LLM effectively removed unnecessary scenes from the input videos, which is important to reduce the complexity of the decision-making process in the task planning for the robot. The success rate of creating a graph representation from the output of the LLM is also very high, showing the LLM followed the provided instructions accurately. In a particular failure case, the LLM created an object node with the type "Dish", which refers to the final cooked food, which is not in the predefined possible object nodes types (ingredients, container, tool or machine). We believe this type of issue could have been caused by a lack of diversity in examples for few-shot prompting, and could be addressed by incorporating additional examples.

The success rate of target object node state estimation from the images was 86\%. The main causes of failure were twofold: (1) the presence of cooking actions in the videos that, while related to the cooking process, were outside the predefined set of allowed actions, and (2) variations in object naming due to paraphrasing.

\begin{table}[bp]
\centering
\caption{Success Rate of Target Object Node Estimation from Videos}
\begin{tabular}{|l|c|}
\hline
\textbf{Task}                     & \textbf{Success Rate} \\ \hline
Removal of unnecessary segments           & 94          \\ \hline
Valid graphs from LLM output            & 98         \\ \hline
Target object node state estimation  & 86          \\ \hline
\end{tabular}
\label{tab:estimationsuccess}
\end{table}

\begin{figure}[tbp]
  \begin{center}
    \includegraphics[width=\linewidth]{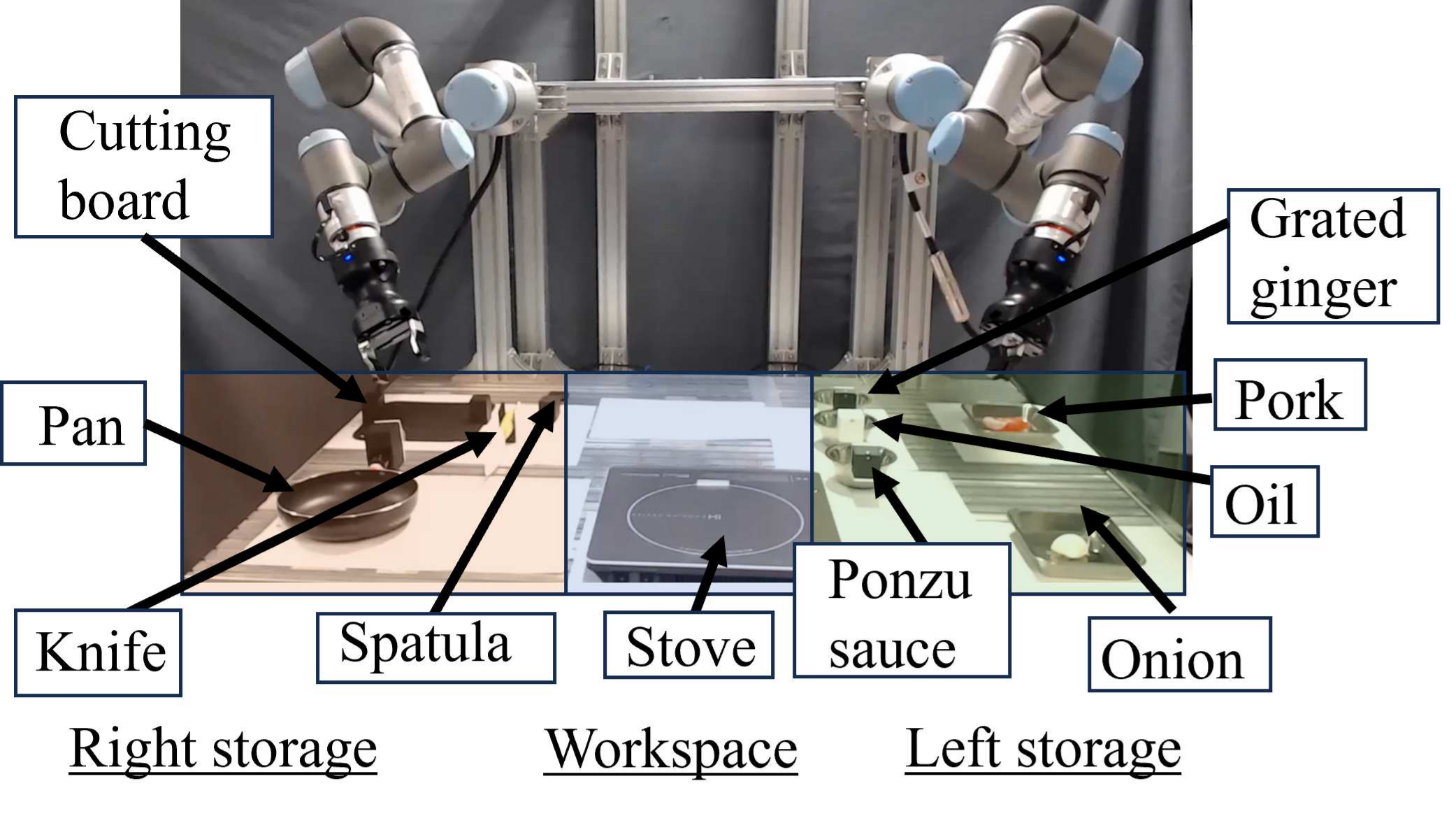}
    \caption{Real-world experimental environment setup for the cooking task with two UR3e robots. The objects are placed in pre-defined areas to ensure all possible required motions are feasible.}
    \label{fig:environment}
  \end{center}
\end{figure}

\begin{figure*}[t] % * をつけると2段組みを超える
    \centering
    \includegraphics[width=\textwidth]{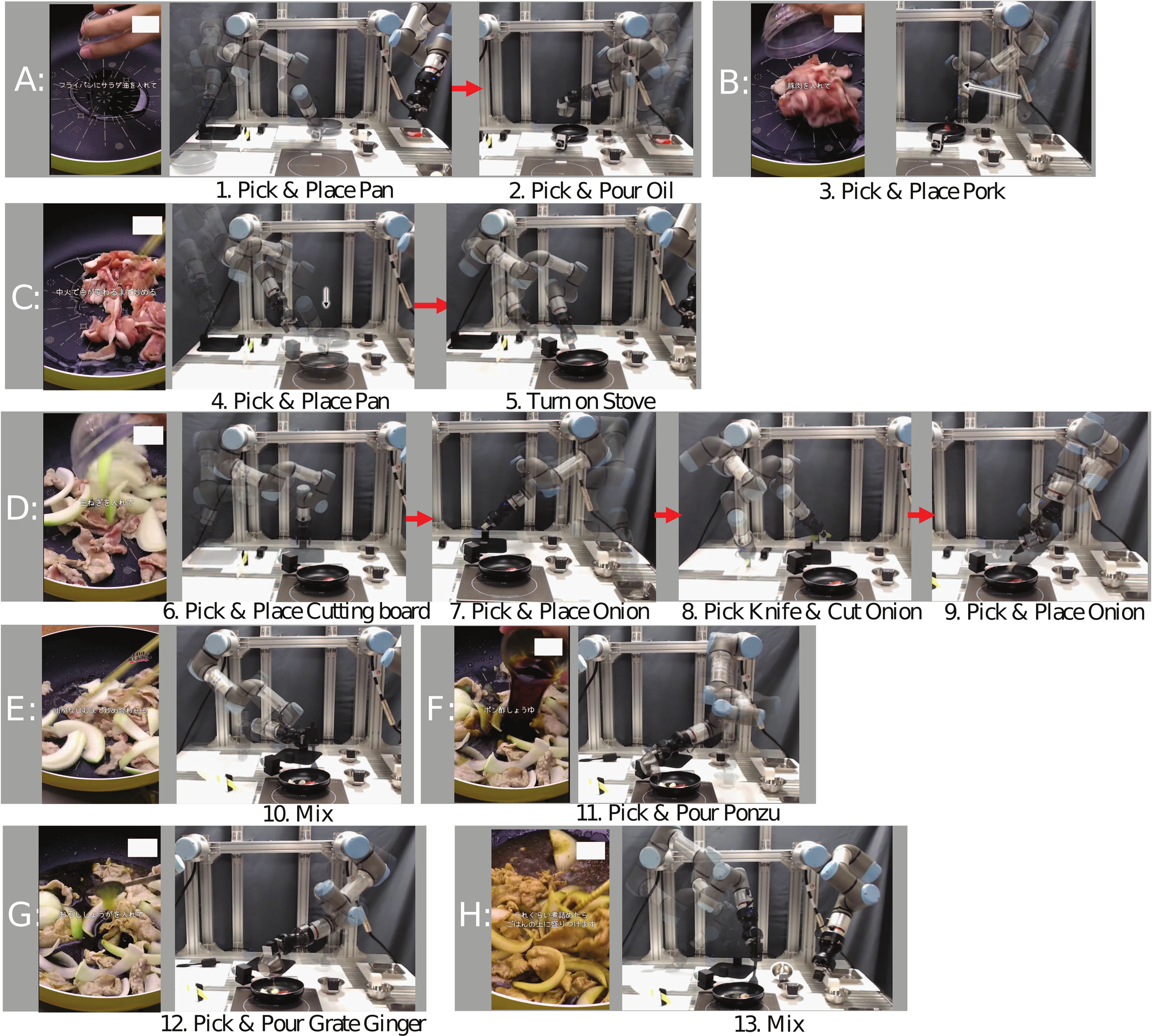} % 画像幅をページ全体に
    \caption{Results for the gyudon (beef bowl) recipe. The system correctly identified key steps from the video, including a omitted step (D. Chopping the onions), to successfully complete the recipe.}
    \label{fig:result}
\end{figure*}

\subsubsection{Generation of Task-sequence graph}

% new
From the 10 videos used for the target object node estimation, we select five cooking recipes to evaluate the full cooking task planning for the robot. We utilize the results of the estimated target object nodes, with slight corrections to three of them to ensure that all estimated target object nodes are feasible, focusing only on the success rate of the full task planning step\cite{YouTube1},\cite{YouTube2}, \cite{YouTube3},\cite{YouTube4}, \cite{YouTube5}. .

The environment object nodes are manually created, with cooking utensils placed in "Right storage" and ingredients placed in "Left storage". We compare the success rate of generating a feasible task plan for the full recipe with our method against a baseline approach, where we only use the results of the LLM with few-shot prompting without the task graph structure validation.

The results are presented in Table \ref{tab:recipe-completion}. The baseline approach could only successfully generate the full task-sequence graph for one recipe. Our proposed method successfully generated task-sequence graphs for four out of the five recipes, demonstrating the effectiveness of our design for generating valid and consistent plans for complex and long-horizon tasks.

The failure case for our method occurred when the LLM, without explicit instructions, determined chopped ingredients on the cutting board should be temporarily placed elsewhere. As a result, it incorrectly places them in a bowl that is later necessary for seasoning preparation, leading to task failure. This shows a limitation of the proposed method: as the validation is performed sequentially and per task plan, once a action node is added to the graph, it cannot be retroactively modified, preventing correction when errors arise in future tasks. This is a point for future improvement, which would further the robustness of the proposed system.

\begin{table}[h!]
\centering
\caption{Comparison of completion rate of task-sequence graph}
\label{tab:recipe-completion}
\begin{tabular}{|l|c|}
\hline
\textbf{Method} & \textbf{Success rate} \\ \hline
Few-shot      & 1/5                      \\ \hline
Ours          & 4/5                          \\ \hline
\end{tabular}
\end{table}

\subsection{Experiment Using Real Robot}
\subsubsection{Experimental environment}

%new
To showcase the proposed method applied in a real robot setup, the cooking recipe for preparing gyudon (beef bowl) is used\cite{YouTube6}. This recipe includes all motion types: "Pick", "Place", "Pour", "Cut", "Mix" and "Cook". However, in the input video, the "Cut" action is omitted, so the LLM must correctly predict the target object node states and infer the missing action. In addition, the cooking video has a title scene and a introduction sequence that shows the completed dish. The LLM also needs ignore such segments and correctly classify and identify the key segments to the cooking task.

The object placement in the real environment is illustrated in Fig. \ref{fig:environment}. Cooking utensils are placed in the "Right storage", while ingredients are placed in the "Left storage". The "Workspace" contains only a stove. The initial placement of all objects is designed to prevent collisions and ensure that Pick motions can be generated successfully. Therefore, the initial placement does not prevent the execution of any Pick motions.

%old
%The cooking recipe used in this study is for preparing gyudon (beef bowl). This recipe includes all motion types: Pick, Place, Pour, Cut, Mix, and Cook. However, in the video, the Cut action is omitted. As a result, the LLM must estimate the target object node states and infer the missing action.
%Additionally, the cooking video contains a title scene and an introduction showcasing the completed dish. Therefore, the LLM must classify and identify the key scenes relevant to task execution.
%The object placement in the real environment is illustrated in Fig. \ref{fig:environment}. Cooking utensils are placed in the "Right storage", while ingredients are placed in the "Left storage". The "Workspace" contains only a stove. The initial placement of all objects is designed to prevent collisions and ensure that Pick motions can be generated successfully. Therefore, the initial placement does not prevent the execution of Pick motions.

%料理レシピは牛丼を作成するレシピを使用する．このレシピは，”Pick”, ”Place”, ”Pour”, ”Cut”, ”Mix”,”Cook”のすべてのモーションが含まれており，動画内ではCutの動作が省略されている．そのため，LLMによる目標オブジェクトノードの状態の推定と，不足動作の補完が必要になる．また料理動画では動画のタイトルや，完成した料理の紹介シーンが含まれており，LLMが重要なシーンを分類する必要がある．実環境でのオブジェクトの配置を，図\ref{fig:environment}に示す．調理器具はRight storageに設置しており，食材はLeft storageに設置されている．また，Workspaceにはコンロだけが設置されている．オブジェクトの初期配置はすべてのオブジェクトが衝突なく，Pickモーションを生成できる座標に配置している．そのため，初期配置によりPickモーションを生成できないことはない．

\subsubsection{Experimental result}

%new
From the input video, it is preprocessed into 13 scenes. The LLM then determines that 8 of them are necessary for the full cooking task, generating the corresponding target object nodes. The LLM successfully eliminated all unnecessary actions, such as the introduction segment, during the target object node estimation step. In the task planning phase, the system successfully generated a full task graph after three rounds of re-planning. The results of the motion planning process are illustrated in Fig. \ref{fig:result}. The video for the full task is available in the supplementary material.

For each recipe scene, the robot manipulated the ingredients to reach the required target object node states. In Fig. \ref{fig:result}D, since the environment did not initially contain chopped onions, the system generated a Cut action using a knife and cutting board before proceeding with the action of placing the onions into the frying pan.
These results demonstrate that the proposed method effectively supplements missing actions while generating feasible task plans.

%old
%In this experiment, the LLM determined that 8 out of 13 scenes in the recipe video were necessary for task execution, generating the corresponding target object nodes. The LLM successfully eliminated unnecessary actions during the target object node estimation process.
%In the task planning phase, the system successfully generated a task graph after three rounds of re-planning. The results of the motion planning process are illustrated in Fig. \ref{fig:result}.
%For each recipe image, the robot manipulated the ingredients to achieve the required target object node states. In Fig. \ref{fig:result}D, since the environment did not initially contain chopped onions, the system generated a Cut action using a knife and cutting board before proceeding with the action of placing the onions into the frying pan.
%These results demonstrate that the proposed method effectively supplements missing actions while generating feasible task plans.
%本実験では，レシピ動画の13個のシーンに対して，8個のシーンをLLMが必要と判断し，目標オブジェクトノードが生成された．LLMによる目標オブジェクトノードの状態推定は，不要な動作を除去することに成功した．また，作業計画では３回の再計画を経て作業グラフの生成に成功した．
%動作計画の結果を図\ref{fig:result}に示す．それぞれのレシピ画像に対して，ロボットにより食材を操作し，目標オブジェクトノードの状態を達成していっている．図\ref{fig:result}Dでは，切られた玉ねぎが環境になかったので，包丁とまな板を用いて切る動作が生成され，その後フライパンに入れる動作が生成されている．以上の結果から，不足する動作を補完しながら作業計画が可能であることが示された．

\section{Conclusions}
%new
We proposed a method to generate a cooking task plan suited for a robot from online cooking videos with subtitles. The proposed method uses an LLM to reason about the content of the video and key state changes of the ingredients during the cooking task. To ensure logical correctness in the proposed task plan, we combine it with a FOON structure, which can also handle the differences between the local environment and the observed environment in the video by using two types of modified functional units: environment object nodes, representing the states of the objects in the local environment, and the target object nodes, representing the states of the objects at the end of each task. This structure can easily detect errors in the task plans, allowing to re-planning until a feasible plan is generated.

The evaluation results show that the proposed system can effectively estimate the relevant segments of a cooking task, and the state of the ingredients at the end of each segment, with a 86\% accuracy across 10 different recipes. Using the correct target node states for all cooking tasks for five recipes, we can successfully generate full task graphs for the local environment for four of them with our approach, whereas if we only use the LLM with few-shot prompting, the full task graph for only one recipe is correctly generated.

A limitation of our proposed method is that the task validation is performed per action node, and it cannot be modified after it is added to the task graph. This prevents adjustment of previous actions if they interfere with future steps, remaining a challenge for future work. Another challenge to be addressed in future work is the error recovery in real-world environments. Currently, it is assumed that the action will succeed if the requirements for the action are met, not taking into account possible errors in the object position estimation and grasp failures. An online system that can update object information in real-time coupled with dynamic re-planning is a key area for future improvement.

\addtolength{\textheight}{-12cm}   % This command serves to balance the column lengths

%%%%%%%%%%%%%%%%%%%%%%%%%%%%%%%%%%%%%%%%%%%%%%%%%%%%%%%%%%%%%%%%%%%%%%%%%%%%%%%%

%%%%%%%%%%%%%%%%%%%%%%%%%%%%%%%%%%%%%%%%%%%%%%%%%%%%%%%%%%%%%%%%%%%%%%%%%%%%%%%%

%%%%%%%%%%%%%%%%%%%%%%%%%%%%%%%%%%%%%%%%%%%%%%%%%%%%%%%%%%%%%%%%%%%%%%%%%%%%%%%%
%\section*{APPENDIX}


%\section*{ACKNOWLEDGMENT}

%This work was supported by the New Energy and Industrial Technology Development Organization (NEDO) under Project JPNP20006

\end{document}